# Preliminary Results of Neuromorphic Controller Design and a Parkinson's Disease Dataset Building for Closed-Loop Deep Brain Stimulation

Ananna Biswas
Department of Electrical and Computer Engineering,
Michigan Tech
Houghton, 49931, USA
anannab@mtu.edu

Hongyu An
Department of Electrical and Computer Engineering,
Michigan Tech
Houghton, 49931, USA
hongyua@mtu.edu

**Abstract—Parkinson's Disease afflicts millions of individuals globally. Emerging as a promising brain rehabilitation therapy for Parkinson's Disease, Closed-Loop Deep Brain Stimulation (CL-DBS) aims to alleviate motor symptoms. The CL-DBS system comprises an implanted battery-powered medical device in the chest that sends stimulation signals to the brains of patients. These electrical stimulation signals are delivered to targeted brain regions via electrodes, with the magnitude of stimuli adjustable. However, current CL-DBS systems utilize energy-inefficient approaches, including reinforcement learning, fuzzy inference, and field-programmable gate array (FPGA), among others. These approaches make the traditional CL-DBS system impractical for implanted and wearable medical devices. This research proposes a novel neuromorphic approach that builds upon Leaky Integrate and Fire neuron (LIF) controllers to adjust the magnitude of DBS electric signals according to the various severities of PD patients. Our neuromorphic controllers, on-off LIF controller, and dual LIF controller, successfully reduced the power consumption of CL-DBS systems by 19% and 56%, respectively. Meanwhile, the suppression efficiency increased by 4.7% and 6.77%. Additionally, to address the data scarcity of Parkinson's Disease symptoms, we built Parkinson's Disease datasets that include the raw neural activities from the subthalamic nucleus at beta oscillations, which are typical physiological biomarkers for Parkinson's Disease.**

## I   INTRODUCTION

Each year, millions of people worldwide are diagnosed with Parkinson's Disease (PD) [1]. Although medications are available for PD treatment, their effectiveness diminishes over time due to drug resistance. Parkinson's patients in later stages must therefore be treated with larger doses of medication, leading to adverse side effects such as depression and speech disorders [1]. Deep Brain Stimulation (DBS) alleviates PD symptoms by delivering electric pulses constantly through the implanted electrode. The electrodes are implanted into the brain through the small hole in the skull. The stimulation signals are generated and modulated by an electronic device within the chest of the patients. The current DBS device provides rigid stimulation signals regardless of the patient's clinical state, leading to side effects. Continuous stimulation signals quickly deplete the DBS device's battery [2]. Hence, an emerging DBS system, namely the Closed-Loop DBS (CL-DBS) system, is proposed to address this issue. CL-DBS systems deliver optimized stimulus impulses based on the different PD symptoms. Beta oscillations (13 to 30 Hz) in the subthalamic nucleus (STN) are typically used as one of the pathological biomarkers for PD symptoms. While the CL-DBS system holds great promise for optimizing therapeutic outcomes in PD patients, one significant challenge lies in its power consumption. Unlike traditional open-loop DBS (OL-DBS) systems, which deliver stimulation continuously or in pre-defined patterns, CL-DBS systems require real-time monitoring of physiological signals and frequent adjustments in stimulation parameters. As a result, CL-DBS systems often consume more power due to the continuous processing of neural signals and the need for rapid computational feedback loops. Current CL-DBS systems are implemented with computationally expensive algorithms and hardware platforms, including reinforcement learning [2, 3], fuzzy inference [4], field-programmable gate array (FPGA) [2], and Artificial Neural Networks (ANNs) [5]. However, these algorithmic approaches are energy-inefficient, making them unsuitable for implanted medical devices.

In this paper, we introduce a novel neuromorphic CL-DBS system using the Leaky Integrate and Fire neuron (LIF) model as stimulation signal controllers. In addition, to address the increasing demands on data for neuromorphic approaches, we built a Parkinson's disease dataset including the raw and beta bandwidth neural activities at the STN region. our contribution is summarized as:

1) Building a dataset of Parkinson's disease based on computational models that include beta oscillation signals in STN and Globus Pallidus internus (GPi) as electrophysiological biomarkers.

2) Several LIF-based controllers for a neuromorphic CL-DBS system are proposed and designed to adjust the magnitude of DBS stimulation signals. Our neuromorphic controllers, the on-off LIF controller, and the dual LIF controller, successfully reduce the power consumption of CL-DBS systems by 19% and 56%, respectively. Meanwhile, the suppression efficiency is increased by 4.7% and 6.77%.

## II  Background of Closed-Loop Deep Brain Stimulation for Parkinson's Disease

Figure 1 (a) illustrates current DBS systems that deliver fixed electric square waveform pulses to specific brain regions to alleviate motor symptoms of PD patients. These regions are typical deep brain structures located within the basal ganglia. These nuclei in basal ganglia are important components of the circuitry that controls motor and movement activities. Both the STN and GPi are primary targets for the DBS system in the treatment of movement disorders. An insulated extension wire connecting the electrode to the implantable pulse generator (IPG) is inserted beneath the patient's skin. The IPG is then subcutaneously placed, usually in the upper chest region. The electrodes for DBS systems are connected to IPG via a tiny, insulated wire that is inserted via a small hole in the skull. Several movement disorders can be treated by DBS, including Parkinson's disease, dystonia, essential tremor [6].

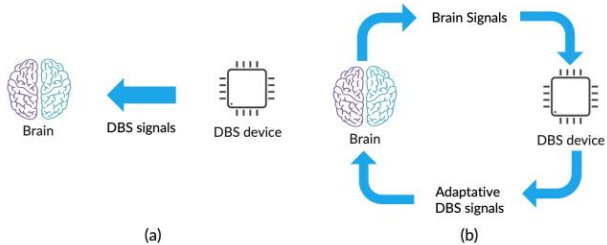

Figure 1: (a) Open-Loop DBS; (b) Closed-Loop DBS.

These conventional DBS systems operate in a one-directional and open-loop manner, delivering stimulation continuously according to pre-defined parameters without adapting to changes in the patient's PD symptom severity or physiological state. As a result, OL-DBS systems have the potential to lead to issues such as battery drainage and serious side effects [7-12].

To address these challenges of simple OL-DBS systems, in recent years, a new approach of adding a new monitor line from the brain of patients to the DBS system is proposed, named as CL-DBS system. Figure 1 illustrates the difference between an OL-DBS and a CL-DBS system. CL-DBS system incorporates real-time feedback from physiological signals or biomarkers to dynamically adjust stimulation parameters in response to changes in the patient's neural activity or symptom severity. CL-DBS is an adaptive system that continuously analyzes the symptoms and indicators of PD and then generates appropriate stimulus signals.

Aiming for stimulus parameters to be updated automatically without user inference, CL-DBS systems require a dependable control mechanism. Making the necessary adjustments to the stimulus signals and performing routine tests for signs of PD are crucial. Additionally, it is imperative to take the required steps to prevent the onset of Parkinson's disease symptoms. Consequently, it is also necessary to build a PD symptom detector and controller that is rapid and energy efficient. The detector must be loaded with complex datasets so that the controller can correctly carry out additional analysis and generate the optimized stimulus signals. The development of CL-DBS systems for PD represents a significant step towards personalized and adaptive treatment strategies tailored to each patient's unique disease progression and symptomatology. By leveraging advances in neuromodulation technology, computational modeling, and signal processing techniques, researchers aim to enhance the precision, effectiveness, and long-term outcomes of DBS therapy for PD patients.

## III  Building a Dataset of Parkinson's Disease using a Computational Model

The data scarcity poses a severe issue for the neuromorphic community when applying Spiking Neural Networks (SNNs) and neuromorphic algorithms to medical applications. In order to address this challenge, we are building a novel dataset of PD biomarkers using a computational model [13].

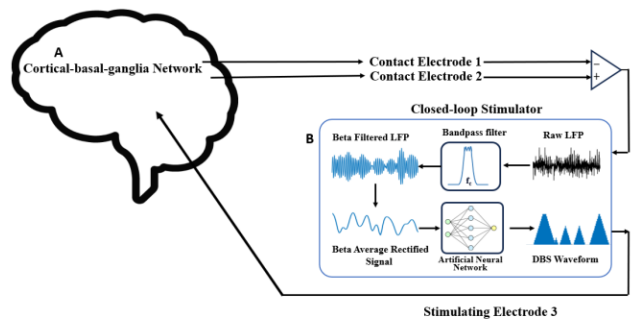

Figure 2: Diagram of Closed-loop DBS System: (A) Network diagram of cortical basal ganglia neuron populations; (B) Diagram of the closed-loop stimulator [13].

The computational model for building the PD dataset is illustrated in Figure 2 [13]. The computation model consists of an extracellular DBS electric field and simulation of the local field potentials (LFP) at STN that is formed between the cortex, basal ganglia, and thalamus [14].

The data is generated from the cortico-basal ganglia computational model, represented as raw local field potentials (LFP) [13]. Local Field Potential refers to the electrical activity recorded from a small group of neurons in the brain. Unlike single-neuron recordings, which focus on the activity of individual neurons, LFP recordings capture the combined activity of nearby neurons. LFP recordings are usually obtained using electrodes implanted in the brain tissue. In the computational model used for building the PD dataset, these electrodes detect the electrical fluctuations generated by the synchronized activity of a population of neurons between the cortex, STN, and thalamus.

The main components of the model are interneurons and cortical neurons of the cortex, STN, globus pallidus externa (GPe), globus pallidus interna (GPe), and thalamus neurons. Cortical pyramidal neurons are simulated using conductance-based biophysical models enabling extracellular DBS electric field to cortical axons. AMPA and GABA imply excitatory synapses and inhibitory synapses respectively. A total of six hundred STN, GPe, GPi, thalamic, cortical interneuron, and cortical pyramidal neurons are connected through these excitatory and inhibitory synapses which are illustrated in Figure 3. The connectivity pattern between neurons in the cortico-basal ganglia network is random. Each of the STN neurons receives 5 inhibitory inputs from GPe neurons and excitatory inputs from five cortical neurons [15]. Each globus pallidus externus (GPe) neuron is subjected to inhibitory input from one striatal neuron and one other GPe neuron while receiving excitatory input from two subthalamic nucleus (STN) neurons. Conversely, each globus pallidus internus (GPi) neuron receives excitatory input from a single STN neuron and inhibitory input from a single GPe neuron. Thalamic neurons encounter inhibitory input from a GPi neuron. Cortical neurons are stimulated by excitatory input from one thalamic neuron and concurrently inhibited by input from ten interneurons. In turn, interneurons are activated by excitatory input from ten cortical neurons [16].

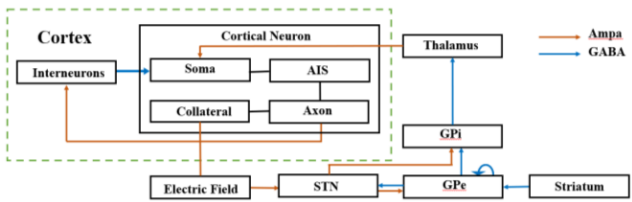

Figure 3: Diagram of cortical basal ganglia network [13].

The cortex is comprised of interneurons and cortical pyramidal neurons. The cortical neuron model consists of soma, axon initial segment (AIS), main axon, and axon collateral. These cortical neuron soma and interneuron models are generated based on regular spiking models. Subthalamic Nucleus includes a leak, sodium, three potassium, two calcium ionic currents, and an intracellular bias current for setting the neuron firing rate. STN plays a vital role in generating bursting activity during Parkinson's disease.

The models for both globus pallidus externus (GPe) and internus (GPi) neurons consist of leak, sodium, two potassium, and two calcium ionic currents, alongside an intracellular bias current that regulates the neuron firing rates. In the case of GPe neurons, an additional intracellular current is introduced to replicate DBS application, with the assumption that a proportionate number of GPe neurons are stimulated as compared to extracellularly stimulated cortical neurons during DBS. Thalamic neurons are also modeled similarly, though one calcium and one potassium current are excluded. The synaptic input from the striatum to GPe neurons is modeled as a collection of Poisson-distributed spike trains operating at a frequency of 3 Hz. The acquired raw LFP is shown in the following Figure 4.

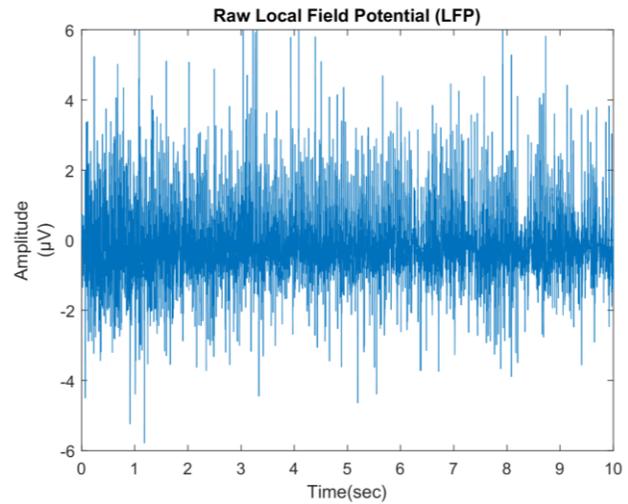

Figure 4: Raw local field potential generated from the neuron population between cortex, STN, and thalamus.

These raw signals of LFP are recorded by the contact electrodes 1 and 2 followed as shown in Figure 2. This is estimated as the summation of the extracellular potentials due to the spatially distributed synaptic currents across the STN population. A bandpass filter is applied to acquire the beta-band filtered LFP. The average rectified value (ARV) of the beta-band LFP is calculated by full-wave rectifying the filtered LFP signal using a fourth-order Chebyshev band-pass filter with an 8 Hz bandwidth, centered about the peak in the LFP power spectrum. The acquired Beta Average rectified signal (ARV) can be seen in Figure 5.

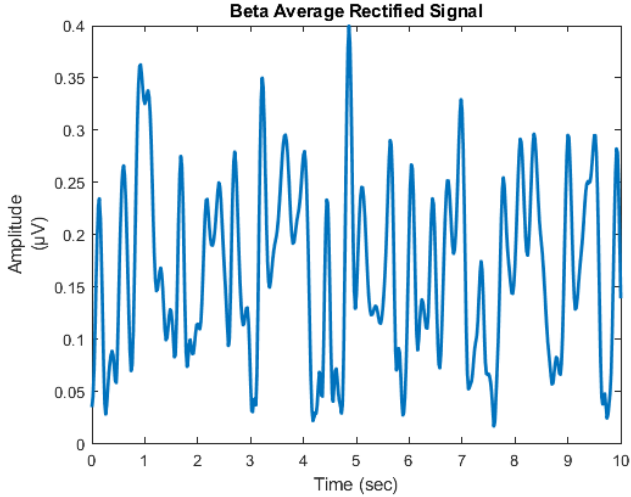

Figure 5: Beta Average Rectified Signal.

IV LEAKY-AND-FIRE NEURON CONTROLLER DESIGN FOR A NEUROMORPHIC CLOSED-LOOP DEEP BRAIN STIMULATION SYSTEM

Two LIF neuromorphic controllers for neuromorphic CL-DBS systems are proposed and designed, namely on-off LIF and dual LIF respectively. The design objective of these LIF controllers is to suppress the power density at beta oscillations from the STN region to a specific target value by adjusting input DBS currents. The LIF neuron models are used for implementing these neuromorphic controllers. In the on-off LIF controller, we set a target value and utilize the LIF neuron model to adjust the DBS current. The maximal DBS current is 3 mA and the minimal DBS current is 0 mA. Specifically, if Beta ARV is larger than the target value, DBS current ($I_{DBS}$) increments, while if Beta ARV is smaller than the target value, DBS current ($I_{DBS}$) keep constant. When the membrane potential of the LIF model crosses the threshold voltage $V_{th}$, the neuron fires and the membrane potential is reset to $V_{leak}$. For the on-off LIF controller, beta ARV is compared to the membrane potential, and the target value is compared to the threshold voltage. When Beta ARV is greater than the target, the DBS current increases according to the Eq. 1 and 2:

$$\frac{dv}{dt} = \frac{\{-(Beta\ ARV - b_{target}) + RI\}}{\tau_m}, \quad (1)$$

$$I_{DBS} = \frac{dv}{dt}/R, \quad (2)$$

where $\tau_m$ is the membrane time constant, $R$ is the membrane resistance, and $I(t)$ is the input current to the neuron. The specific values of these parameters are listed in Table 1.

TABLE 1: PARAMETERS OF ON-OFF LIF DBS CONTROLLER

| $V_m(t)$ | $V_{th}$ | $\tau_m$ | $R$ | $I$ | $b_{target}$ |
|---|---|---|---|---|---|
| Measured Beta ARV | $b_{target}$ | 5 | 0.5 Ω | 5 mA | 0.104 μV |

In contrast, when the Beta ARV is lower than the target, the DBS current (mA) is constant. The simulation results are demonstrated in Figure 6. The beta band has been extracted from the raw Local Field Potential (LFP). The beta average rectified value (ARV) of the beta-band LFP is calculated by full-wave rectifying the filtered LFP signal using a fourth-order Chebyshev band-pass filter with an 8 Hz bandwidth, centered about the peak in the LFP power spectrum. In the time interval from 11 sec to 12 sec, the DBS current is constant but before 11 sec there is an upward trend as in that time the beta ARV is greater than the target value. The maximum and minimum range of DBS current is 3 mA and 0 mA respectively.

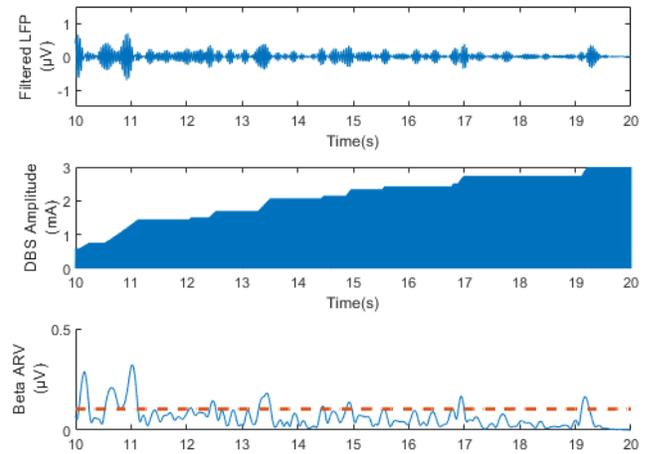

Figure 6: Adaptive DBS current of on-off LIF controller.

The dual threshold LIF model utilizes the same equations (Eq. 1 and Eq. 2) to adjust the DBS current. However, for the dual threshold, there are two target values. The upper target value is set as 0.104 μV and the lower target is set as 0.05207 μV as shown in TABLE 2. Consequently, when Beta ARV is larger than the upper target, the DBS current increments, while when Beta ARV is smaller than the lower target, the DBS current decreases. If the Beta ARV is in the range between the upper target and lower target, the DBS current remains constant. Figure 7 illustrates the dual threshold LIF controller output.

TABLE 2: PARAMETERS OF DUAL LIF DBS CONTROLLER.

| $V_m(t)$ | $V_{th}$ | $\tau_m$ | $R$ | $I$ | Targets (μV) |
|---|---|---|---|---|---|
| Measured Beta ARV | Targets | 5 | 0.5 Ω | 5 mA | $t_{up}$= 0.104<br>$t_{low}$= 0.05207 |

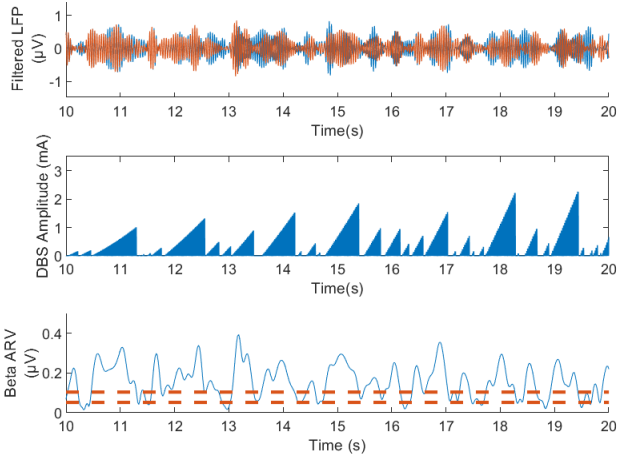

Figure 7: Adaptive DBS current of dual LIF controller.

These LIF based neuromorphic controllers are assessed using three critical parameters, which are mean squared error (MSE), power consumption, and suppression efficiency.

The mean squared error (MSE) measures the capability of controllers to detect the beta target level. The MSE is defined quantitatively using the following equations:

$$e(t) = \frac{b_{measured} - b_{target}}{b_{target}} \quad (3)$$

$$MSE = \frac{1}{T_{sim}} \int_0^{T_{sim}} e(t)^2 \, dt, \quad (4)$$

where $T_{sim}$ is the simulation duration (30 seconds), $e(t)$ is the normalized error signal between the measured LFP beta ARV and the target value. The MSEs for all the controllers is calculated as a percentage of the MSEs measured in each respective scenario when DBS is off. This value is considered the baseline for comparisons among different controllers.

Figure 8 represents the MSE of different CL-DBS controllers, including open-loop controller, on-off LIF controller and dual LIF controller. The error is 100% when the DBS stimulus is not provided. In this circumstance, the beta ARV signal reflects the pathological beta activity oscillation. Hence, the error is maximal in no DBS stimulus scenario. The MSEs of other controllers in Figure 8 are the normalized signal between the beta ARV signal and the target beta signal. Lower MSE values are indicative of enhanced controller performance, reflecting superior alignment between the actual and desired signals. The less error signifies the smaller beta ARV signal. Moreover, the beta ARV signal is modulated by the applied DBS current in the cortical-basal-ganglia network. Therefore, it is necessary to consider the magnitude of DBS current required to achieve the desired control over beta oscillation. Figure 8 illustrates the on-off LIF controller (11%) has less error than the Dual LIF controller (30%).

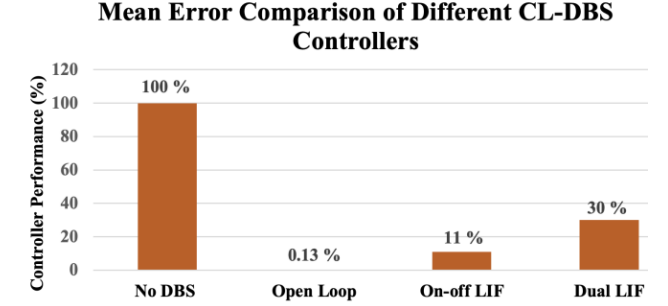

Figure 8: Comparison of mean squared error (MSE) among different CL-DBS controllers.

The power consumption of the controllers is measured as follows:

$$\text{Power Consumption} = \frac{1}{T_{sim}} \int_0^{T_{sim}} Z_E(t) I_{DBS}(t)^2 \, dt, \quad (5)$$

where the $Z_E$ is the electrode impedance (0.5 k $\Omega$), $I_{DBS}$ is the DBS current, and $T_{sim}$ is the simulation duration (30 seconds). Figure 9 represents the power consumption among the different controllers. The power consumption of all controllers is normalized with respect to the Open-loop controller using 2.5 mA DBS current. Notably, the Open-loop controller exhibits 100% power consumption, the highest among all controllers, due to its continuous DBS current application. This constant DBS current is administered to modulate the beta ARV signal in the cortical-basal-ganglia network.

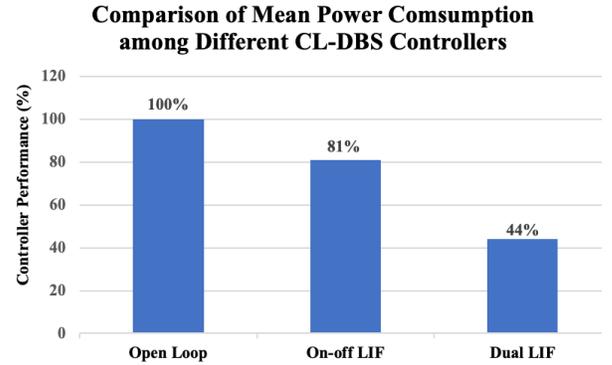

Figure 9: Comparison of the mean power consumption among different CL-DBS controllers.

Based on the delivered DBS current, $I_{DBS}$, the power consumption of neuromorphic controllers is calculated. The power consumption indicates the amount of DBS current required to regulate the beta band oscillation of the patient's brain. Lower power consumption signifies a more efficient controller. The On-off LIF controller consumes 81 percent of the power, which is 0.54 times greater than the power consumed by the Dual LIF controller. Among all three controllers, the Dual LIF controller exhibits the lowest power consumption (44%), making it the most efficient controller.

The last assessment parameter of controllers in a CL-DBS system is the suppression efficiency. This efficiency is quantified as the percentage of beta suppression per unit of power consumed, with units %/µW. The controller suppression efficiency is defined as:

$$\text{Suppression Efficiency} = 100 \times \frac{1 - \frac{1}{T_{sim}} \int_0^{T_{sim}} \frac{b_{DBSOff}(t) - b_{controller}(t)}{b_{DBSOff}(t)} dt}{\text{Power Consumption}}, \quad (6)$$

where $b_{DBSOFF}$ is the beta ARV signal measured in the simulation when DBS is off, $b_{controller}$ is the beta ARV signal measured with the controller simulation active, and the power consumption is the power used by the controller in the simulation, as defined in Eq. (5). The suppression efficiency calculates how efficiently each controller can suppress the beta band oscillation of the pathological brain. There is excessive beta oscillation observed in the basal ganglia of Parkinson's patients [17]. Different types of controllers reduce this oscillation based on the target beta level. The higher the suppression efficiency, the more efficient the controller is at regulating the beta ARV signal.

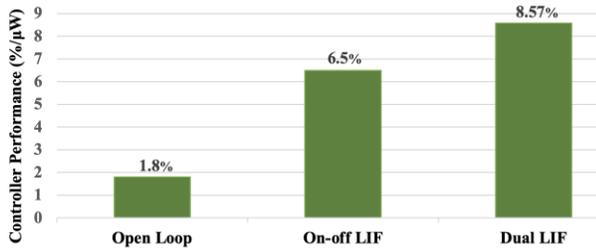

Figure 10: Comparison of suppression efficiency among different CL-DBS controllers.

The suppression efficiency comparison among different controllers is illustrated in Figure 10. The suppression efficiency demonstrates a reverse relationship with the power consumption of each controller. As the Open-loop controller consumes the highest power among all controllers, it exhibits the lowest efficiency at 1.8%/µW. Conversely, the On-off LIF controller (6.5%/µW) demonstrates greater efficiency than the Open-loop controller, with its efficiency being 3.61 times higher. Overall, the Dual LIF controller emerges as the most efficient, boasting a suppression efficiency of 8.57%/µW.

## V CONCLUSION

This study introduces a novel neuromorphic approach, leveraging Leaky Integrate and Fire (LIF) neuron controllers, to tailor traditional CL-DBS electric signals according to varying PD severities. Our proposed controllers, the on-off LIF controller, and dual LIF controller, significantly reduce CL-DBS system power consumption by 19% and 56%, respectively, while increasing suppression efficiency by 4.7% and 6.77%. Moreover, to address the data scarcity of PD symptoms, we curated PD datasets containing raw neural activities from the subthalamic nucleus (STN) at beta oscillations, crucial physiological biomarkers for PD diagnosis and treatment. In addition, a novel database is generated for the neuromorphic community to study, analyze, and design customized CL-DBS system and is published at https://github.com/Brain-Inspired-AI-Lab/Parkinson-Electrophysiological-Signal-Dataset-PESD. The dataset includes the neural signals with different PD states.

## ACKNOWLEDGEMENT

This work was supported by the program: Disability and Rehabilitation Engineering (DARE) of the National Science Foundation under Award Number 2301589. In addition, the authors deeply appreciate Dr. Kuba Orłowski and Dr. Madeleine Lowery from University College Dulin for generously sharing their knowledge on Parkinson's disease and closed-loop Deep Brain Stimulation.